\documentclass[runningheads]{llncs}
\usepackage[T1]{fontenc}
\usepackage[misc]{ifsym}
\newcommand{\corr}{(\Letter)}
\usepackage{graphicx}
\usepackage{xcolor}
\usepackage{floatrow}
\usepackage[colorlinks,
            linkcolor={black},
            citecolor={blue},
            urlcolor={blue},
            pdfauthor={No Author Given},
            bookmarks=false,
            pdfsubject={ART: Actually Robust Training},
            pdfkeywords={accuracy, confidence intervals, confidence bounds, cross-validation, bootstrapping, holdout}
            ]{hyperref}

\begin{document}

\title{\textsc{ART}: Actually Robust Training}
\toctitle{\textsc{ART}: Actually Robust Training}
\titlerunning{\textsc{ART}: Actually Robust Training}

\author{Sebastian Chwilczyński\orcidID{0009-0004-1894-9131} \and
Kacper Trębacz\orcidID{0009-0006-5615-5935} \and
Karol Cyganik\orcidID{0009-0001-7935-3719} \and
Mateusz Małecki\orcidID{0009-0007-2080-3096} \and
Dariusz Brzezinski\orcidID{0000-0001-9723-525X} \corr}
\tocauthor{Sebastian Chwilczyński, Kacper Trębacz, Karol Cyganik, Mateusz Małecki, Dariusz Brzezinski}
\authorrunning{S. Chwilczynski et al.}

\institute{Institute of Computing Science, Poznan University of Technology, Poland\\\email{dariusz.brzezinski@cs.put.poznan.pl}}
\maketitle
\begin{abstract}
Current interest in deep learning captures the attention of many programmers and researchers. Unfortunately, the lack of a unified schema for developing deep learning models results in methodological inconsistencies, unclear documentation, and problems with reproducibility. Some guidelines have been proposed, yet currently, they lack practical implementations. Furthermore, neural network training often takes on the form of trial and error, lacking a structured and thoughtful process. To alleviate these issues, in this paper, we introduce \textsc{Art}, a Python library designed to help automatically impose rules and standards while developing deep learning pipelines. \textsc{Art} divides model development into a series of smaller steps of increasing complexity, each concluded with a validation check improving the interpretability and robustness of the process. The current version of \textsc{Art} comes equipped with nine predefined steps inspired by Andrej Karpathy’s Recipe for Training Neural Networks, a visualization dashboard, and integration with loggers such as Neptune. The code related to this paper is available at: \url{https://github.com/SebChw/Actually-Robust-Training}.

\end{abstract}

\section{Introduction}
The process of developing deep learning models is complex and abstract. This is due to several factors, including the vast number of configuration choices during neural network training and countless opportunities for introducing hard-to-find bugs, like wrong data normalization. As the field is evolving rapidly, the development of deep learning pipelines is neither standardized nor well-documented, resulting in extensive time spent understanding someone's code, finding errors, and reproducing their results. 

Existing deep learning frameworks either focus on improving computational efficiency~\cite{pytorch}, conciseness of code~\cite{lightning}, automatic hyperparameter tuning~\cite{automl}, or parallelization~\cite{memento}. Guidelines and best practices, on the other hand, usually reside in books or articles~\cite{karpathy2019recipe} and remain detached from implementation tools. Even though many deep learning toolkits are available~\cite{frameworks}, to the best of our knowledge, there exists no unified framework that strongly enforces best practices and incremental design as part of model development process.

To fill this gap, we have developed \textsc{Art} (\textit{Actually Robust Training}). \textsc{Art} is a Python library inspired by Andrej Karpathy’s Recipe for Training Neural Networks~\cite{karpathy2019recipe} that promotes good practices such as iterative design and testing. Its purpose is to make the development of deep learning models more manageable and reproducible by creating a unified interface. To achieve this, \textsc{Art} contributes PyTorch~\cite{pytorch} and Lightning~\cite{lightning} companion classes that:
\begin{itemize}
    \item standardize the development of deep learning projects;
    \item provide debugging, experiment tracking, and visualization utilities;
    \item include pre-defined best practice steps and tutorials;
    \item offer the option to create custom steps, validation checks, and decorators.
\end{itemize}
In the following sections, we present the main components of \textsc{Art}.

\section{The \textsc{Art} package}
\textsc{Art} was designed based on two principles: \textit{“Build from simple to complex”} and \textit{“Verify the success of every action with an experiment”}. To implement these principles, we use the concept of a \textit{project} that divides deep learning model development into a sequence of \textit{steps}, each equipped with some form of validation (\textit{check}).

\begin{figure}
    \centering
    \includegraphics[width=0.8\textwidth]{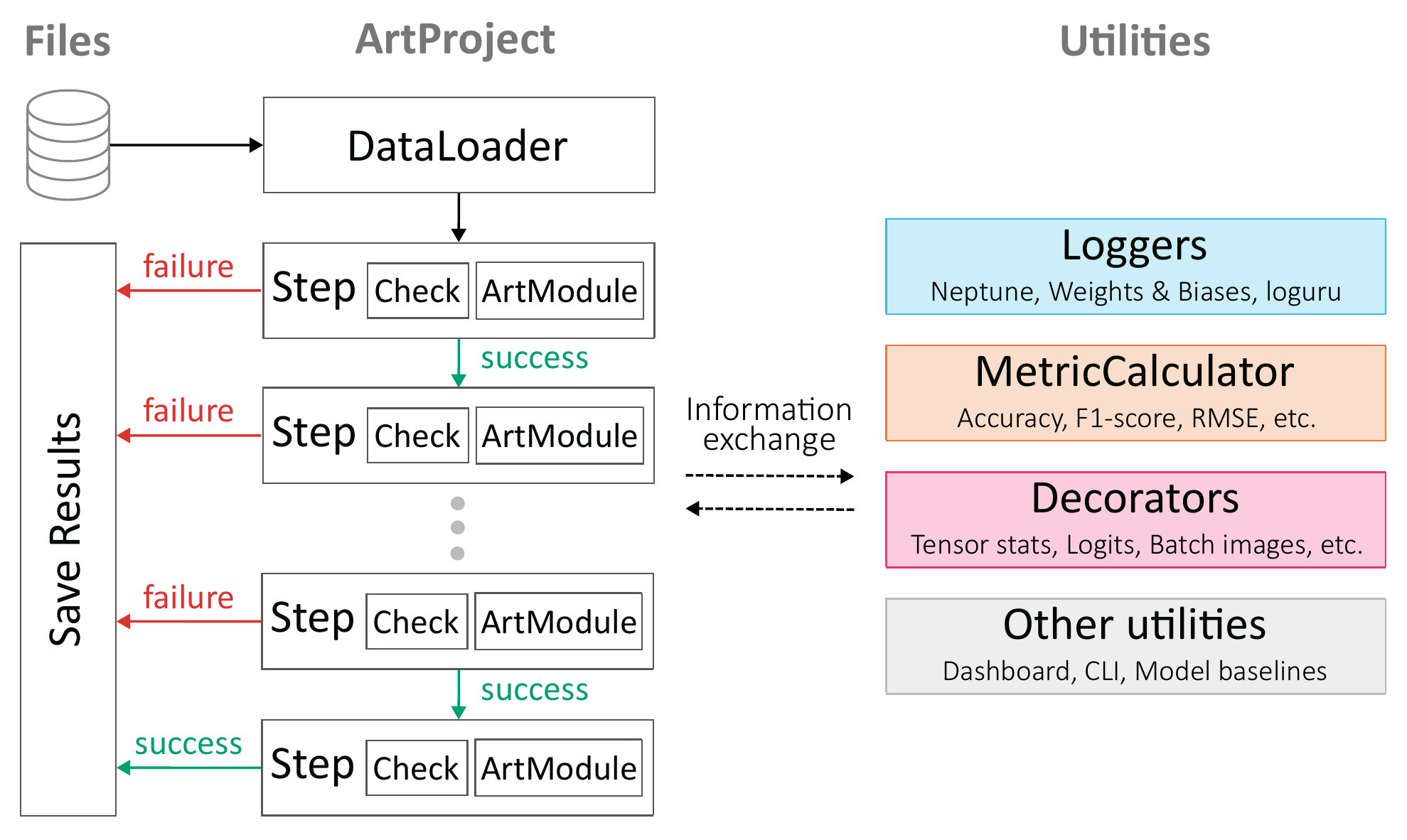}
    \caption{Schematic representation of \textsc{Art}'s components. Deep learning projects are stored as ArtProjects, which consist of steps. Each step validates a version of the deep learning model (ArtModule) using a check function. If the check is successful, the project moves on to the next step, following the \textit{build from simple to complex} principle. \textsc{Art} also offers utilities for measuring, logging, and visualizing the whole process.}
    \label{fig:components}
\end{figure}

More precisely, every research project in \textsc{Art} is stored in an \texttt{ArtProject} container, which consists of a PyTorch Lightning \texttt{DataModule} and a series of \texttt{Steps} combined with \texttt{Checks}. Each \texttt{Step} is a modular and independent experiment pushing the project forward. A \texttt{Check} is a boolean function validating the success of the performed step. Moreover, every element of \textsc{Art} uses a \texttt{MetricCalculator}, which calculates all losses and metrics throughout every project experiment, ensuring comparability between steps. This approach provides modularity and interpretability of the development process. A schematic showing the relation between different \textsc{Art} components is presented in Figure~\ref{fig:components}.

The idea of dividing the development process into steps follows Karapthy's Recipe for Training Neural Networks~\cite{karpathy2019recipe}. \textsc{Art} comes with several pre-implemented steps such as: \textit{Analyze data}, \textit{Check loss on init}, \textit{Overfit one batch}, \textit{Regularize}, and \textit{Transfer learning}. Most importantly, \textsc{Art} allows developers to create custom steps, tailored to their needs or required by company procedures.

By dividing projects into steps, \textsc{Art} aims to help reduce the number of bugs introduced during model development. Errors in deep neural network implementations are especially annoying since they rarely cause an exception but rather reveal themselves only by a deterioration of the model's predictive performance. Detecting such exception-less bugs can make or break a deep learning project. To aid developers in this endeavor, \textsc{Art} comes equipped with three complementary debugging functionalities: \textit{decorators}, \textit{loggers}, and a \textit{dashboard}. 

\begin{figure}
    \centering
    \includegraphics[width=0.95\linewidth]{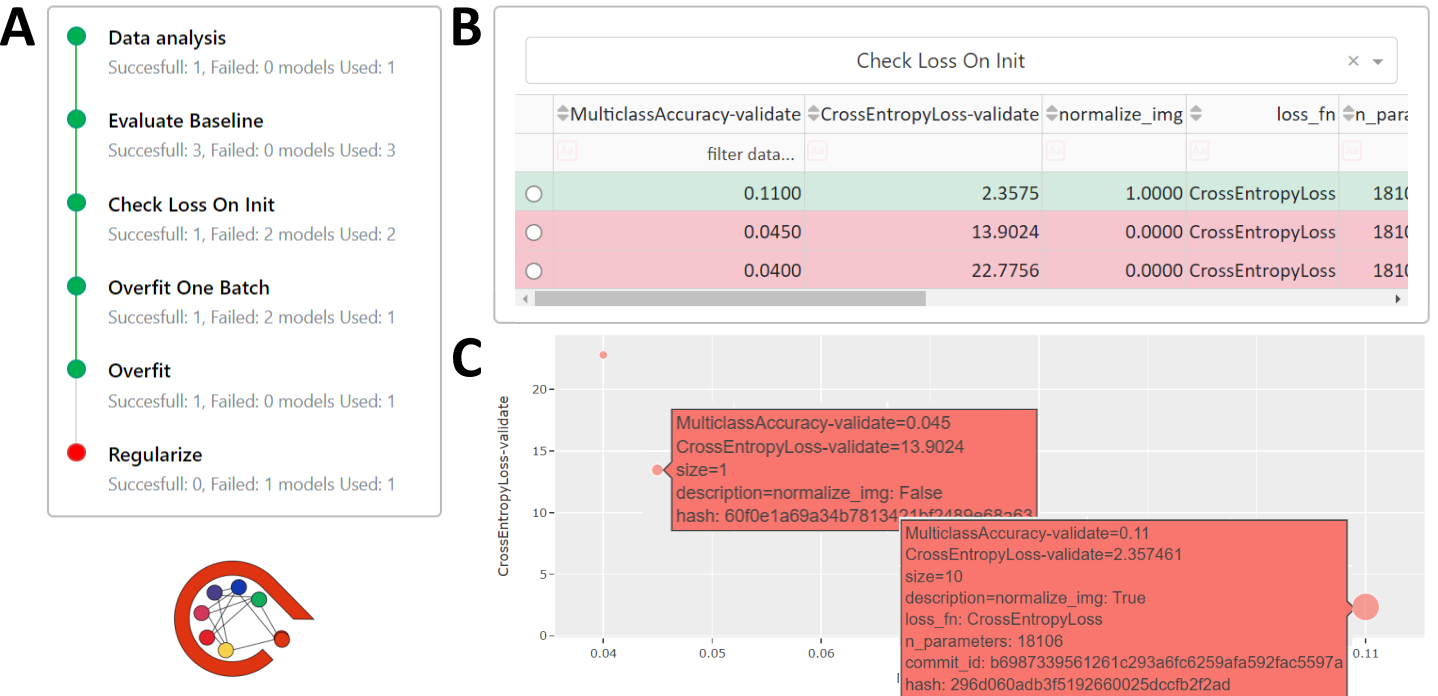}
    \caption{\textsc{Art}'s Dashboard. (A) List of steps and their statuses. (B) Results of different model versions. (C) Plot for comparing model versions.}
    \label{fig:dashboard}
\end{figure}

A \texttt{Decorator} is an easy-to-implement class wrapping any part of the deep learning training pipeline. Decorators allow developers to add extra debugging functionality by just decorating Python objects and not explicitly modifying the model's code. For example, using the \texttt{BatchSaver} decorator, one can save batch input data (e.g., a set of images) to a separate folder for visual inspection. Additionally, every information an \texttt{ArtProject} gathers can be tracked using third-party experiment \texttt{Loggers}. Currently, \textsc{Art} offers integration with experiment tracking servers such as Neptune~\cite{neptune} and Weights \& Biases~\cite{wandb}, as well as filesystem logging with loguru~\cite{loguru}. 
Finally, the whole training process of each run can be visualized on a Dashboard (Figure~\ref{fig:dashboard}). Everyone can track the project's progress by analyzing the differences between consecutive steps, \textsc{Art}'s unique feature. Importantly, \textsc{Art} detects changes in the code of \texttt{ArtModules} and \texttt{Dataloaders} and prompts information that the \texttt{Step}'s status may require re-running.

\textsc{Art}'s source code and documentation are available on GitHub\footnote{\url{https://github.com/SebChw/Actually-Robust-Training}} and an introductory video can be viewed at \url{https://youtu.be/3KH14mjhwEo}. 

\section{Target users and applications}
\textsc{Art} has been designed with four user groups in mind: \textit{students}, \textit{educators}, \textit{researchers}, and \textit{developers}.

\textbf{Students.} Modern deep learning frameworks provide very extensive high-level APIs. For people just starting with deep learning, it may be difficult to find causes of problems when something does not work. When creating tutorials with \textsc{Art}, one can easily adjust their difficulty by selecting appropriate steps. Additionally, using \textsc{Art}, students can learn good practices from early beginnings.

\textbf{Educators.} Checking assignments is a time-consuming task. With \textsc{Art}, educators can create assignments from existing blocks and have validation for free. It is also easier to guide a student when one knows the step a mistake was made. Finally, \textsc{Art} comes with a set of tutorials, which we hope will expand very soon.

\textbf{Researchers.} Lack of reproducibility is a common problem in deep learning research. \textsc{Art} directly tries to mitigate this issue by imposing standards and logging the learning process results. \textsc{Art} also makes it easy to spot model changes thanks to code hashing and model comparisons.

\textbf{Developers.} By following the \textit{“Verify the success of every action”} principle, \textsc{Art} is a great tool for model testing and continuous integration, which in the long run may bring big benefits. Moreover, with \textsc{Art}'s modularity, developers can easily implement steps that adhere to company procedures.

\section{Conclusions}
In this paper, we introduced \textsc{Art}, a Python library that helps students, researchers, and engineers follow best practices while developing deep learning models. \textsc{Art}'s steps, checks, decorators, and loggers allow for easy progress tracking, model debugging, and reproducibility. To the best of our knowledge, \textsc{Art} is the first framework focused on deep learning best practices. Although the proposed library is currently limited to PyTorch models, it could be extended to other deep learning toolkits in the future. We also envision \textsc{Art} as a community hub where deep learning developers share knowledge through step templates and show how to adhere to standards, such as the upcoming EU AI Act.

\subsubsection*{Acknowledgements}
This research was partly funded by PUT Statutory Funds and the National Science Centre, Poland, grant number 2022/47/D/ST6/01770. For the purpose of Open Access, the author has applied a CC-BY public copyright license to any Author Accepted Manuscript (AAM) version arising from this submission.

\bibliographystyle{splncs04}

\end{document}